\documentclass[conference]{IEEEtran}
\IEEEoverridecommandlockouts
\usepackage{cite}
\usepackage{amsmath,amssymb,amsfonts}
\usepackage{algorithmic}
\usepackage{graphicx}
\usepackage{textcomp}
\usepackage{xcolor}
\usepackage[bookmarks=false]{hyperref}

\def\BibTeX{{\rm B\kern-.05em{\sc i\kern-.025em b}\kern-.08em
    T\kern-.1667em\lower.7ex\hbox{E}\kern-.125emX}}

\usepackage[utf8]{inputenc} 
\usepackage[T1]{fontenc}    
\usepackage{url}            
\usepackage{booktabs}       
\usepackage{amsfonts}       
\usepackage{nicefrac}       
\usepackage{microtype}      

\usepackage{comment}
\usepackage{graphicx} 

\usepackage{soul}

\usepackage{times,epsfig,graphicx,amsmath}
\usepackage{times,graphicx,amsmath,amssymb,booktabs,tabulary,multirow,overpic,xcolor,bbm,amssymb}
\usepackage{caption}
\usepackage{subcaption}
\definecolor{citecolor}{RGB}{34,139,34}

\usepackage[bookmarks=false]{hyperref}

\newcommand{\app}{\raise.17ex\hbox{$\scriptstyle\sim$}}

\newcolumntype{x}[1]{>{\centering\arraybackslash}p{#1pt}}

\newlength\savewidth\newcommand\shline{\noalign{\global\savewidth\arrayrulewidth
  \global\arrayrulewidth 1pt}\hline\noalign{\global\arrayrulewidth\savewidth}}
\newcommand{\tablestyle}[2]{\setlength{\tabcolsep}{#1}\renewcommand{\arraystretch}{#2}\centering\footnotesize}
\makeatletter\renewcommand\paragraph{\@startsection{paragraph}{4}{\z@}
  {.5em \@plus1ex \@minus.2ex}{-.5em}{\normalfont\normalsize\bfseries}}\makeatother

\begin{document}

\title{Efficient Deep Gaussian Process Models for\\ Variable-Sized Inputs}

\author{
\IEEEauthorblockN{Issam H. Laradji\IEEEauthorrefmark{1}\IEEEauthorrefmark{2},
Mark Schmidt\IEEEauthorrefmark{1}, Vladimir Pavlovic\IEEEauthorrefmark{3}\IEEEauthorrefmark{4},
Minyoung Kim\IEEEauthorrefmark{3}\IEEEauthorrefmark{5}}
\IEEEauthorblockA{\IEEEauthorrefmark{2}\textit{Element AI, Montreal, Canada}\\
\IEEEauthorrefmark{1}\textit{Dept. of Computer Science,  University of British Columbia, Vancouver, Canada}\\
\IEEEauthorrefmark{3}\textit{Dept. of Computer Science, Rutgers University, Piscataway, New Jersey, USA}\\
\IEEEauthorrefmark{4}\textit{Samsung AI Center, Cambridge, UK}\\
\IEEEauthorrefmark{5}\textit{Dept. of Electronic Engineering, Seoul Nat'l Univ. of Science \& Technology, Seoul, South Korea}\\
\IEEEauthorrefmark{1}\{issamou, schmidtm\}@cs.ubc.ca,
\IEEEauthorrefmark{3}vladimir@cs.rutgers.edu,
\IEEEauthorrefmark{4}v.pavlovic@samsung.com,
\IEEEauthorrefmark{5}mikim21@gmail.com
}}

\maketitle

\begin{abstract}
Deep Gaussian processes (DGP) have appealing Bayesian properties, can handle variable-sized data, and learn deep features. Their limitation is that they do not scale well with the size of the data. Existing approaches address this using a deep random feature (DRF) expansion model, which makes inference tractable by approximating DGPs. However, DRF is not suitable for variable-sized input data such as trees, graphs, and sequences. We introduce the GP-DRF, a novel Bayesian model with an input layer of GPs, followed by DRF layers. The key advantage is that the combination of GP and DRF leads to a tractable model that can both handle a variable-sized input as well as learn deep long-range dependency structures of the data. We provide a novel efficient method to simultaneously infer the posterior of GP's latent vectors and infer the posterior of DRF's internal weights and random frequencies. Our experiments show that GP-DRF outperforms the standard GP model and DRF model across many datasets. Furthermore, they demonstrate that GP-DRF enables improved uncertainty quantification compared to GP and DRF alone, with respect to a Bhattacharyya distance assessment. Source code is available at { https://github.com/IssamLaradji/GP\_DRF}.

\end{abstract}

\section{Introduction}

Deep neural network (DNN) models have achieved ground-breaking performance in many real-life domains such as computer vision and natural language processing~\cite{goodfellow2016deep}. This is mainly due to their ability to model long-range dependency structures that may reside in the data. However, they do not provide uncertainty quantification, which can be useful in decision making and high risk applications such as medical informatics or autonomous driving~\cite{gal2016uncertainty}. A recent method addresses this limitation using random feature expansion~\cite{rfdnn17}, but is unable to efficiently handle variable-sized data, such as trees~\cite{moschitti2006making}, protein sequences~\cite{lo2000scop}, audio sequences~\cite{kuksa2008kernel}, or graphs~\cite{vishwanathan2010graph}, in an end-to-end fashion. For instance, to predict the chemical properties of variable-sized molecular data~\cite{DuvMacetal15nfp}, a separate feature extraction stage was required in order to construct fixed-sized fingerprint vectors. Such sophisticated feature extraction schemes often require human expertise. In this work, we propose Gaussian Process Deep Random Feature (GP-DRF), a scalable Bayesian method, that addresses the aforementioned limitations.

Bayesian models have received significant attention over the last decade. Gaussian processes (GP) are a family of flexible function distributions that can exploit the kernel trick to avoid dealing with input instances directly, leading to a scalable approach for computing instance similarities, and uncertainties about the latent functions. This is determined by the covariance of the GP model using a kernel function. Hence, the many choices of kernels, such as string kernel functions for sequence classification~\cite{farhan2017efficient}, enable GPs to handle variable-sized data effectively in the Euclidean, non-Euclidean, and RKHS metric spaces. However, GPs are shallow and therefore  unable to benefit from the key properties of DNNs, namely exploiting the deep long-range dependency structures in the data through a long chain of composite operations.

\begin{figure}
\begin{center}
\includegraphics[trim = 0mm 0mm 0mm 0mm, clip, scale=0.42]{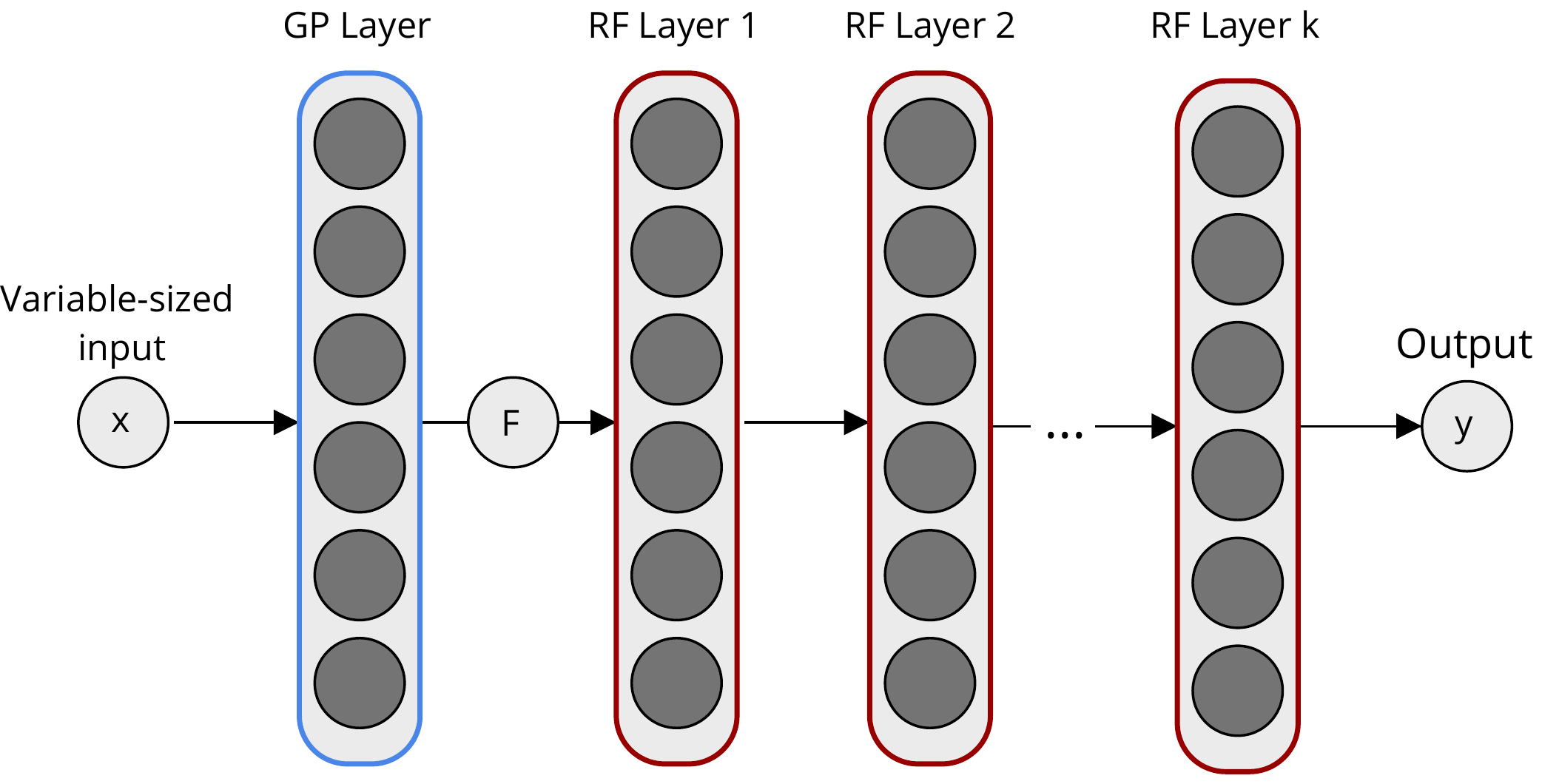} 
\end{center}
\vspace{-0.8em}
\caption{GP-DRF. A layer of Gaussian processes (GP) first maps a possibly variable-sized input $x$ to a fixed-sized latent feature vector $F$. Then, a set of random feature (RF) expansion layers maps those features to a prediction output $y$.}
\vspace{-0.5em}
\label{fig:overview}
\end{figure}

To overcome the aforementioned shortcoming, Damianou {\em et al.}~\cite{deepgp13} proposed deep GP models. The main idea of deep GPs is to replace each layer of linear-to-nonlinear mappings in the DNN by a layer of random (nonlinear) functions sampled from a GP, thus achieving a long chain of composite operations where all the functions involved have their own GP priors. However, a critical weakness of this approach is that it is highly intractable; the functions in one layer take as the input the outputs of the latent functions in the prior layer, implying that the kernels are built on the outputs of other latent functions. As a result, the marginalization of the latent functions in a long chain of compositions becomes computationally expensive. Methods such as the inducing point approach~\cite{quinonero2005unifying} have been proposed to address the scalability issue of deep GPs. However, many parameters still need to be inferred, making the deep GP impractical for many large-scale applications~\cite{rfdnn17}.

Several works address the computational difficulty of the non-parametric kernel machines by transforming them into parametric models using random feature expansion~\cite{rf_fourier, rf_arccos}. Random features are (nonlinear) feature vector representations that approximate, in expectation, the kernel values of the feature vector inner products. For a Gaussian process, the latent function can be expressed as a linear function in the feature space with a Gaussian-priored weight vector, leading to a {\em parametric} Bayesian model. As  a result, the kernel matrices need not be stored or inverted, leading to a dramatically improved computational efficiency.  Recently, Cutajar {\em et al.}~\cite{rfdnn17} proposed a deep random feature (DRF) model as a parametric formulation of the deep GP by approximating its kernel features. Their results showed that DRF can yield significant computational benefits compared to deep GPs, while maintaining comparable performance in their benchmarks. However, DRF is limited only to kernels that are shift-invariant, such as the radial basis (RBF) and arc-cosine kernel functions. Consequently, DRF have difficulty dealing with variable-sized data, in contrast to generic sequence, tree, and graph kernels, which can be leveraged by Gaussian process models. 

The motivation of this work is to build a deep GP model that is not only as scalable as DRF, but can also handle variable-sized data in the same manner as GPs coupled with kernel machines, while achieving comparable performance. To that end, we propose a GP-DRF, which combines a single layer of a GP model with multiple layers of DRF models. As shown in~Fig.\ref{fig:overview}, the GP model represents the first layer that takes arbitrarily shaped data as the input and returns fixed-sized feature vectors of latent functions as the output. The upper DRF model then maps the feature vectors to the prediction space. For the GP-DRF, we propose an efficient variational inference scheme that can handle large-scale data using pseudo-inputs as inducing points, in a fashion similar to Dai~{\em et al.}~\cite{dai2015variational} and Bui~{\em et al.}\cite{bui2016deep}.

We summarize the benefits of GP-DRF as follows:
\begin{enumerate}
\item It can effectively handle variable-sized data by learning sequence kernels using the GP component in its first layer, unlike DRF-only models;
\item 
It is a more accurate representation of a deep GP model than DRF, as GP-DRF 's first layer is the exact non-parametric formulation of the GP model, whereas DRF layers are all approximations to GPs;
\item It is a scalable approximation of deep GPs as the random feature expansion method allows computationally efficient training and inference;
\item The Bayesian nature of our model allows it to estimate uncertainty, which is crucial for many real-world applications; and
\item It outperformed DRF and GPs on several classification, regression, and uncertainty benchmarks.
\end{enumerate}

\section{Background on Random Features in GP and DRF}\label{sec:background}
This section briefly reviews the DRF model~\cite{rfdnn17} which can accurately approximate deep GPs. Since the core idea of the DRF is to model each layer in DNN by the random-feature expansion of the GP, we begin the discussion with random features and its applications in GPs. Note that the DRF model assumes, and is restricted to, fixed-sized inputs, hence we denote $d$ as the dimensionality of the input vector $x$ throughout this section.  

Random features~\cite{rf_fourier,rf_arccos} have a finite dimensional feature vector representation for inputs where the inner product on their feature space equals (approximately and/or in expectation) the value of the kernel function of interest. That is, the aim is to find a $D$-dim feature vector $\phi(x)$ such that for a given kernel $k(x,x')$, we have $\phi(x)^\top \phi(x') \simeq k(x,x')$. For instance, the ARD kernel,
\begin{equation}
k_{ARD}(x,x') = \alpha \exp \Big( -\frac{1}{2}(x-x')^\top \Gamma^{-1} (x-x') \Big),
\label{eq:kernel_ard}
\end{equation}
with parameters $\theta = \{ \alpha,
\Gamma=\textrm{diag}(\gamma_1,\dots,\gamma_d) \}$, 
admits the $D=2M$-dim feature representation: 
\begin{equation}
\begin{split}
\phi_{ARD}(x) = \sqrt{\frac{\alpha}{M}} \ \Big[
  \cos(\omega_{(1)}^\top x), \sin(\omega_{(1)}^\top x), \dots, \\
  \cos(\omega_{(M)}^\top x), \sin(\omega_{(M)}^\top x) \Big]^\top,
\end{split}
\label{eq:rf_ard}
\end{equation}
where $\omega_{(m)}$ for $m=1,\dots,M$ are i.i.d. samples from 
$\mathcal{N}(0, \Gamma^{-1})$~\cite{rf_fourier}. Refer to Cho {\em et al.}~\cite{rf_arccos} for more details about the arc-cosine kernel feature representations. 

We acquire several advantages by having a random feature representation for the kernel. Particularly in the GP, as we need $\textrm{Cov}(f(x),f(x')) = k(x,x')$ for every input pair $(x, x')$, it can be achieved by having a Bayesian {\em parametric} linear model on top of the feature space, namely, 
$f(x) = w^\top \phi(x)$ with prior $w \sim \mathcal{N}(0, I_D)$. This is because 
$\textrm{Cov}(f(x),f(x')) = \textrm{Cov}(w^\top \phi(x), w^\top \phi(x')) = 
\phi(x)^\top \phi(x') = k(x,x')$. Hence, using the random feature expansion method, one can reduce all inference operations of a non-parametric GP to a parametric formulation, which significantly improves computational time as illustrated in Cutajar {\em et al.}~\cite{rfdnn17}. This is because the setup does not require storing training data points nor kernel matrices, thus avoiding the need to perform kernel matrix inversions which are costly. In other words, we have a succinct summary about the posterior information through the finite dimensional $w$, namely $P(w|\mathcal{D})$.


Next, we briefly describe how the DRF model~\cite{rfdnn17} approximates the deep GP model~\cite{deepgp13} using random feature expansion. In the deep GP model, the $l$-th layer ($l=0,\dots,L-1$) taking $h^l \in \mathbb{R}^{d_l}$ as input (by convention, $h^0 = x$ and $h^L$ is the model's final output) is modeled as:
\begin{equation}
\begin{split}
h^{l+1}_j = f^l_j(h^l)\ \ &\textrm{with} \ \ 
f^l_j(\cdot) \sim \mathcal{GP}(k^l(\cdot,\cdot)) 
\ \ \\ &\textrm{for} \ \ j=1,\dots,d_{l+1}.
\end{split}
\label{eq:deepgp_lth}
\end{equation}
Note that the functions within the $l$-th layer  (i.e $f^l_j$ for $j=1,\dots,d_{l+1}$) do not necessarily need to have identical GP prior defined by the kernel function $k^l(\cdot,\cdot)$; each function can have its own prior. In the DRF model, the random feature expansion replaces GP-priored $f^l_j(\cdot)$ in Eq.~\ref{eq:deepgp_lth} by Gaussian-priored linear functions of random features, yielding:
\begin{equation}
\begin{split}
h^{l+1}_j = {w^l_j}^\top \phi^l(h^l) \ \ &\textrm{with} \ \ 
w^l_j \sim \mathcal{N}(0, I_{D_l}) 
\ \ \\ &\textrm{for} \ \ j=1,\dots,d_{l+1},
\end{split}
\label{eq:rfdnn_lth}
\end{equation}
where $\phi^l(h^l)$ is the $D_l$-dim feature vector corresponding to 
$k^l(\cdot,\cdot)$, the kernel in the $l$-th layer. If it is ARD, for instance, 
one can use the form defined in Eq.~\ref{eq:rf_ard} with $x$ replaced by $h^l$. 

That is, the $l$-th layer of the DRF model is, using the vector forms and 
explicitly specifying the dependency of $\phi^l(\cdot)$ on the random spectra, 
can be written as:
\begin{equation}
\begin{split}
h^{l+1} = {W^l}^\top \phi^l(h^l; \Omega^l) \ \ &\textrm{with} \ \ 
W^l \sim \mathcal{N}(0, I)\ \ \\ &\textrm{and} \ \ 
\Omega^l \sim \mathcal{N}(0, \Lambda^l).
\end{split}
\label{eq:rfdnn_lth_vec}
\end{equation}
$W^l$ is a $(D_l \times d_{l+1})$ matrix where $w^l_j$ represents its $j$-th 
column, $\Omega^l$ denotes all the random spectra $\omega$'s in the random features 
$\phi^l(\cdot)$ (such as those in Eq.~\ref{eq:rf_ard}), and $\Lambda^l$ defines the parameters of the density from which the random spectra are sampled (for example, $\Lambda^l=\Gamma^{-1}$ 
for the ARD kernel) where we assumed zero-mean Gaussian\footnote{Although there 
exist random features based on non-Gaussian samples, we confine all our derivations 
to the Gaussian density due to simplicity and popularity. 
Nonetheless, this can be extended to non-Gaussian densities where sampling is easy, and evaluating the corresponding Gaussian-expected log-density and its gradient is easy to carry out, at least approximately.}.  

Now, cascading Eq.~\ref{eq:rfdnn_lth_vec} for $l=0,\dots,L-1$ forms the feed-forward 
function of the DRF, which is denoted as $y = G(x)$. That is,
\begin{equation}
G(x; W, \Omega, \theta_o) = g^{L-1}(\cdots (g^1(g^0(x))) \cdots),
\label{eq:rfdnn_fwd}
\end{equation}
where $h^{l+1} = g^l(h^l; W^l, \Omega^l, \theta^l_o)$ are shorthand for 
Eq.~\ref{eq:rfdnn_lth_vec} where $\theta^l_o$ indicates parameters other than 
$W^l$ and $\Omega^l$ in the $l$-th layer (this includes the output variance parameter 
$\alpha^l$ defined in Eq.~\ref{eq:rf_ard}). We have also denoted $W = \{ W^l \}_{l=0}^{L-1}$ 
(similarly for $\Omega$ and $\theta_o$).

\begin{figure}
\begin{center}
\includegraphics[trim = 0mm 0mm 0mm 0mm, clip, scale=0.425]{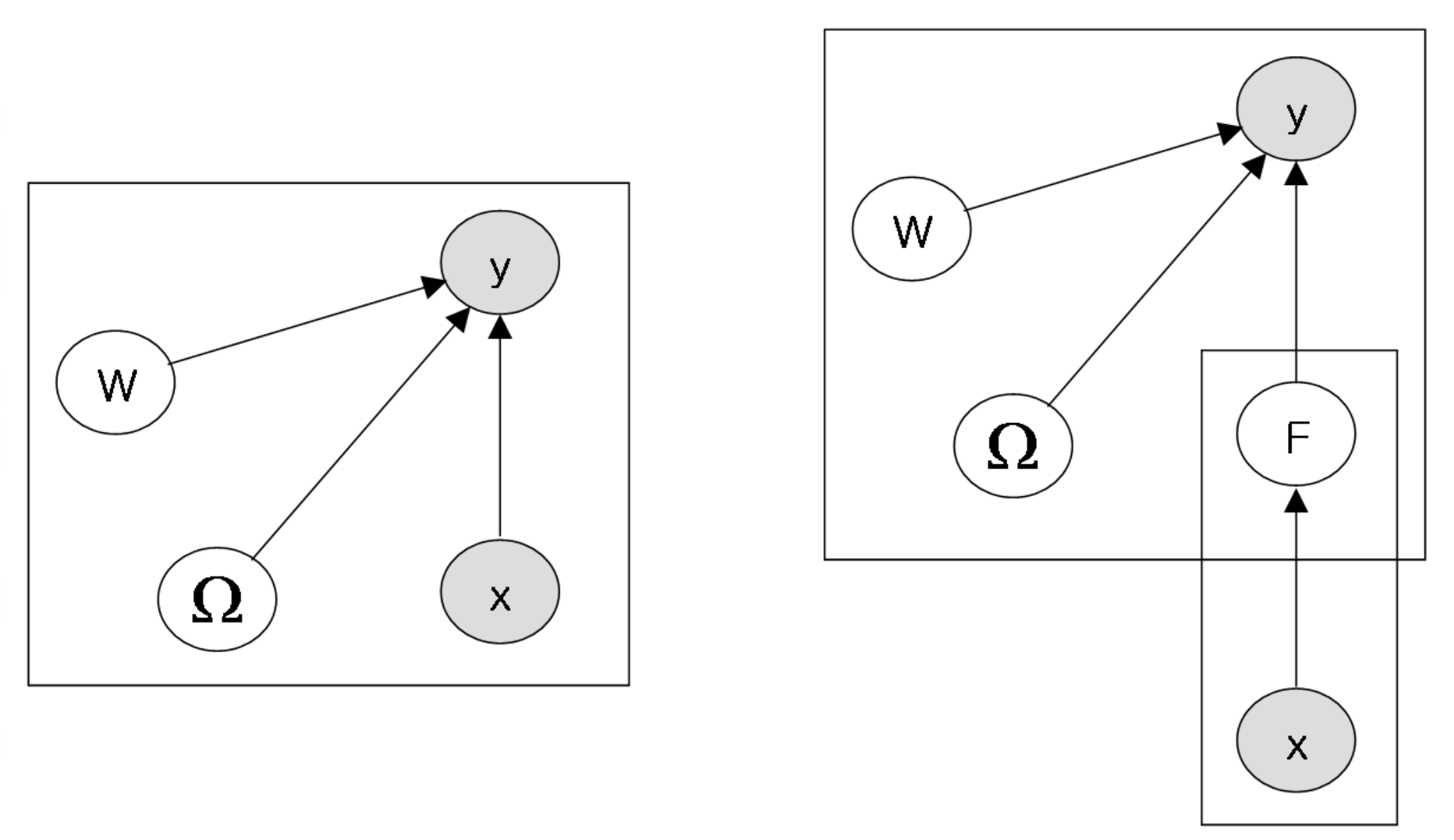}
\end{center}
\vspace{-0.8em}
\caption{Graphical model representations for (Left) DRF and (Right) the proposed model, GP-DRF.}
\label{fig:rfdnn}
\vspace{-0.5em}
\end{figure}

DRF is a deep Bayesian neural network model that addresses the critical drawback of deep GPs by making inference much more scalable using a parametric formulation. However, the random feature expansion method can only be applied to a restricted class of kernel function. First, random feature representations are only known for a limited number of kernel functions such as ARD and arc-cosine. Second, it is not applicable to kernel functions that operate on variable-sized inputs as they are not shift invariant (it is not feasible to define a shift operation for a pair of variable-sized inputs), according to Bochner's theorem~\cite{rudin_book}. This poses a limitation on DRF's ability to deal with sequence data. In the next section, we introduce a novel model that extends DRF that can deal with variable-sized inputs.

\section{Proposed approach}\label{sec:ours}

We propose GP-DRF, a deep Bayesian model that uses GP's kernel machines (which can utilize sequence kernel functions for variable-sized inputs) in conjunction with the deep architecture of the DRF model. As shown in Fig.~\ref{fig:rfdnn},  the GP layer is placed at the bottom which takes possibly variable-sized input data $x$ as input and returns a vector of latent functions $F$ as output. $F$ is then fed into the upper DRF model as input which maps it to the prediction space.  

In the next section, we provide a detailed description of the semi-parametric model that is GP-DRF. Further, we propose an efficient variational inference method for computing the posterior of GP's latent vector, used as input to the DRF model, in conjunction with the posterior of the internal weights and random spectra of the DRF model. To make the GP layer computationally efficient, we use the inducing point method~\cite{dai2015variational} in our implementation.

\subsection{Model Architecture}\label{sec:model}

The bottom layer of GP-DRF is a GP whose latent functions operate on 
(possibly variable-sized) input $x$, and returns an output vector that is fed into the 
upper DRF model as input. More specifically, we consider $d_0$ latent 
functions $\{f_j(\cdot)\}_{j=1}^{d_0}$ (so, $d_0$ becomes the input dimension 
of the DRF), and each latent function is drawn from $\mathcal{GP}(k_j(\cdot,\cdot))$ 
independently from one another.  

We are given $N$ training instances $\mathcal{D}=\{(x_n, y_n)\}_{n=1}^N$ 
where each $x_n$ is an (sequence) input and $y_n$ is the corresponding target 
value (For example, a discrete class label for classification or real-valued for regression). Often, the input data alone are separately denoted by $X = \{x_1,\dots,x_N\}$, and $Y =\{y_1,\dots,y_N\}$. As the model contains the non-parametric component (the bottom GP layer), we need to maintain the outputs of the latent functions as random variables. Formally, we denote them by the $(N \times d_0)$ matrix, 
$F = [F_1, \dots, F_N]^\top$ where its $n$-th row contains the GP's output 
vector for $x_n$, denoted by (using subscript)  $F_n = [f_1(x_n), \dots, f_{d_0}(x_n)]^\top$. The $j$-th column of $F$  consists of the outputs of the $j$-th function over all input instances,  denoted by (using superscript), $F^j = [f_j(x_1), \dots, f_j(x_N)]^\top$  for $j=1,\dots,d_0$. From the aforementioned independent GP prior assumption, $F$ is distributed as a Gaussian distribution factorized over $j$:
\begin{equation}
P(F) = \prod_{j=1}^{d_0} \mathcal{N}(F^j; 0, K_j),
\label{eq:our_P(F)}
\end{equation}
where $K_j$ is the $(N \times N)$ kernel matrix extracted from $X$ using the kernel 
function $k_j(\cdot,\cdot)$ for $f_j(\cdot)$. 

For each instance $n$, the output $F_n$ from the GP layer serves as input to 
the DRF model, resulting in the final output $G(F_n; W, \Omega, \theta_o)$ 
by following Eq.~\ref{eq:rfdnn_fwd}. We link this output to the target $y_n$ 
by a likelihood model. The likelihood model can be chosen according to the prediction task (some examples are logistic or probit model for class-labeled $y$, and Gaussian for real-valued $y$). We denote the likelihood model as: 
\begin{equation}
P(y_n|G(F_n; W, \Omega, \theta_o), \theta_l),
\label{eq:our_lik}
\end{equation}
where $\theta_l$ stands for the parameters of the likelihood model (for instance, 
the weight vector in a logistic model or the noise variance in a Gaussian). 
As is common in practice, we assume the data instances are i.i.d., which lets 
the total likelihood be a product of Eq.~\ref{eq:our_lik} over $n=1,\dots,N$.

Placing the priors on $W$ and $\Omega$ in the upper DRF model together, 
the full joint likelihood of our GP-DRF model can be written as follows: 
\begin{equation}
\begin{split}
P(Y, W, \Omega, F | X, \Theta) = 
&P(F|\theta_k) P(W) P(\Omega|\Lambda) \\
&\prod_{n=1}^N P(y_n|G(F_n; W, \Omega, \theta_o), \theta_l),
\label{eq:our_full_lik}
\end{split}
\end{equation}
where (i) $\theta_k$ indicates the parameters of all the kernel functions 
$k_j(\cdot,\cdot)$ in Eq.~\ref{eq:our_P(F)};
(ii) $P(W) = \prod_l \mathcal{N}(W^l; 0, I)$; 
(iii) $P(\Omega|\Lambda) = \prod_l \mathcal{N}(\Omega^l; 0, \Lambda^l)$; and 
(iv) $\Theta = \{ \theta_k, \theta_l, \theta_o, \Lambda \}$ which represents all 
the parameters of the GP-DRF model.

\subsection{Variational Inference}\label{sec:inference}

In this section, we describe the inference formulation for the posterior distribution of the underlying latent variables of the GP-DRF model, specifically
\begin{equation}
P(F , W, \Omega | X, Y, \Theta).
\label{eq:post_orig}
\end{equation}
A main benefit of our approach is that from Eq.~\ref{eq:post_orig}, we can quantify the uncertainty about not only the parameters of the deep model ($W$ and $\Omega$), but also the inputs ($F$) to the deep model. 

To perform inference, we opt for the popular variational inference 
method that uses pseudo inputs~\cite{titsias09,dezfouli15}. This is computationally feasible for large-scale data as the complexity grows linearly with $N$. Furthermore,  this allows for mini-batch type variational optimization since the log of Eq.~\ref{eq:post_orig} admits the form of {\em summation} of the log-likelihoods over instances. We introduce $M (\ll N)$ as pseudo inputs, denoted by 
$\overline{X}=\{\overline{x}_1, \dots, \overline{x}_M \}$. The pseudo inputs 
can be either selected randomly from $X$, or chosen as representatives by performing
clustering on $X$. Note that clustering variable-sized data is feasible as sequence kernels can operate directly on points in $X$. The latent function vectors on $\overline{X}$ are denoted as $\overline{F}$, 
similarly as we defined $F$.

Next, we introduce the variational density $q(\cdot)$ that approximates 
Eq.~\ref{eq:post_orig}. In defining $q(\cdot)$, we assume fully factorized Gaussians 
for $W$ and $\Omega$ for computational simplicity. 
For $F$, we force the conditional density $q(F|\overline{F})$ to coincide with 
the prior $P(F|\overline{F})$, which is crucial to have some difficult terms 
canceled out, making the inference scalable~\cite{titsias09}. In essence, the 
variational density is defined as:
\begin{equation}
q(W, \Omega, F| \Psi) = q(W| \Psi_W) \ q(\Omega| \Psi_\Omega)  
    \int P(F|\overline{F}) \ q(\overline{F}| \Psi_F) \ d\overline{F},
\label{eq:inf_q}
\end{equation}
where 
\begin{eqnarray}
q(W| \Psi_W) &=&    \label{eq:inf_qW}
    \prod_{l,i,j} \mathcal{N}(w^l_{i,j}; m^l_{i,j}, (s^l_{i,j})^2) \\ 
q(\Omega| \Psi_\Omega) &=&    \label{eq:inf_qOmega}
    \prod_{l,i,j} \mathcal{N}(\omega^l_{i,j}; \eta^l_{i,j}, (\beta^l_{i,j})^2) \\
q(\overline{F}| \Psi_F) &=&    \label{eq:inf_qF}
    \prod_{j=1}^{d_0} \mathcal{N}(\overline{F}^j; \mu_j, \Sigma_j),
\label{eq:inf_q_each}
\end{eqnarray}
where the notations are described as follows.
(i) $w^l_{i,j}$ (scalar) is  the $(i,j)$-element of $W^l$, and all the variational 
parameters for $q(W)$ are denoted as $\Psi_W=\{ (m^l_{i,j},s^l_{i,j}) \}$ 
(similarly for $\omega^l_{i,j}$ and $\Psi_\Omega$), 
(ii) $\mu_j$ and $\Sigma_j$ are $M$-dim mean vector and $(M \times M)$ full 
covariance matrix for Gaussian $q(\overline{F}^j)$, 
where $\Psi_F = \{ (\mu_j, \Sigma_j) \}$, and 
(iii) $\Psi=\{\Psi_W, \Psi_\Omega, \Psi_F\}$ indicates the entire variational 
parameters. 

The following inequality, derived from the KL divergence between $q(\cdot)$ 
and the posterior Eq.~\ref{eq:post_orig}, provides the lower bound of the 
log-evidence.
\begin{equation}
\log P(Y| X, \overline{X}, \Theta) \geq \textrm{ELBO}(\Psi, \Theta),
\label{eq:ineq_elbo}
\end{equation}
where the evidence lower-bound (ELBO) is defined as:
\begin{eqnarray}
\textrm{ELBO}(\Psi, \Theta) &=&  
\sum_{n=1}^N \mathbb{E}_q [ \log P(y_n|G(F_n; W, \Omega, \theta_o), \theta_l)]  \nonumber \\ 
&& \ - \ 
  \textrm{KL}(q(W,\Omega,\overline{F})||P(W,\Omega,\overline{F})).  
\label{eq:elbo}
\end{eqnarray}
Since the bounding gap in Eq.~\ref{eq:ineq_elbo} is exactly the KL divergence 
between $q(\cdot)$ and the posterior, increasing $\textrm{ELBO}(\Psi, \Theta)$ 
with respect to $\Psi$ leads to a better variational density, 
whereas increasing with respect to $\Theta$ may improve the data evidence score of the model. Hence, maximizing $\textrm{ELBO}(\Psi, \Theta)$ with respect to both variable sets
can achieve variational inference and  model selection. 

Next we describe how to evaluate the objective $\textrm{ELBO}(\Psi, \Theta)$ 
and its gradient. The second term Eq.~\ref{eq:elbo} is comprised of KL divergences between Gaussians, which admit closed forms and are easy to derive. 
The first term, as briefly mentioned earlier, has the form of a summation over 
the data instances, which can be readily approximated by a mini-batch 
average over a small subset of data (thus scalable to a large dataset via 
stochastic gradient~\cite{ranganath2014black}. Now we explain each individual term $n$ ($=1,\dots,N$), that is,
\begin{equation}
\mathbb{E}_q [ \log P(y_n|G(F_n; W, \Omega, \theta_o), \theta_l)].
\label{eq:ell_n}
\end{equation}
Note that the expectation is with respect to $$q(W, \Omega, F_n) = 
q(W) q(\Omega) q(F_n).$$ For $q(F_n)$, the integration in the third term of 
Eq.~\ref{eq:inf_q} can be done analytically, yielding a Gaussian: 
$q(F_n) = \mathcal{N}(F_n; a_n, B_n)$. 
Specifically, the mean vector $a_n$ is $(d_0 \times 1)$ and the covariance 
matrix $B_n$ is $(d_0 \times d_0)$ diagonal, and their $j$-th elements can 
be written as ($j=1,\dots,d_0$): 
\begin{eqnarray}
{[a_n]}_j &=&    \label{eq:qF_anj}
    \overline{k}_j(x_n)^\top (\overline{K}_j)^{-1} \mu_j, \\
{[B_n]}_{j,j} &=&    \label{eq:qF_Bnj}
    k_j(x_n,x_n) - 
    \overline{k}_j(x_n)^\top (\overline{K}_j)^{-1} \overline{k}_j(x_n) \\&&+  
     \;\overline{k}_j(x_n)^\top (\overline{K}_j)^{-1} \Sigma_j 
        (\overline{K}_j)^{-1} \overline{k}_j(x_n), \ \ \
\end{eqnarray}
where $\overline{k}_j(x_n) = 
[ k_j(x_n,\overline{x}_1), \dots, k_j(x_n,\overline{x}_M) ]^\top$ and 
$\overline{K}_j$ is the $(M \times M)$ kernel matrix for $k_j(\cdot,\cdot)$ 
on pseudo inputs $\overline{X}$.

Although the expectation is taken with respect to the Gaussian distribution, the log-likelihood is a highly complex function of the integration variables $W$, $\Omega$, and $F_n$, and thus it cannot be done analytically. Furthermore, when we take the gradient of Eq.~\ref{eq:ell_n} with respect to $\Psi$ and 
$\Theta$, we should note that the underlying density $q(\cdot)$ is dependent 
on both of these variable sets. 
To overcome this difficulty, we follow the re-parametrized Monte-Carlo 
estimation technique suggested by Kingma {\em et al.}~\cite{autoenc_vb14} for the Bayesian DNN, and also adopted in Cutajar {\em et al.}~\cite{rfdnn17} for the parametric inference of the DRF model. The idea is to re-parametrize the Gaussian integration variables by decomposing them into parameters that we optimize over and random variables that are 
parameter-free. More specifically, we re-write each variable as:
\begin{eqnarray}
w^l_{i,j} &=&    \label{eq:reparam_W}
    m^l_{i,j} + s^l_{i,j} e_{lij}, \ \ \ \ \ \ 
    e_{lij} \sim \mathcal{N}(0,1), \\ 
\omega^l_{i,j} &=&    \label{eq:reparam_Omega}
    \eta^l_{i,j} + \beta^l_{i,j} \tau_{lij}, \ \ \ \ \ \ 
    \tau_{lij} \sim \mathcal{N}(0,1), \\ 
{[F_n]}_j &=&    \label{eq:reparam_F}
    [a_n]_j + {[B_n]}^{1/2}_{j,j} \epsilon_{nj}, \ \ \ \ \ \ 
    \epsilon_{nj} \sim \mathcal{N}(0,I_{d_0}).
\end{eqnarray}
After sampling $S$ sets of independent standard normal random numbers 
$\{ e^{(s)}_{lij}, \tau^{(s)}_{lij}, \epsilon^{(s)}_{nj} \}_{l,i,j,n}$ for 
$s=1,\dots,S$, we plug these into Eq.~\ref{eq:reparam_W}--\ref{eq:reparam_F} 
to get the sample versions of $(W^{(s)}, \Omega^{(s)}, F^{(s)}_n)$, and 
have an unbiased estimate of Eq.~\ref{eq:ell_n}:
\begin{equation}
\frac{1}{S} \sum_{s=1}^S 
    \log P(y_n|G(F^{(s)}_n; W^{(s)}, \Omega^{(s)}, \theta_o), \theta_l).
\label{eq:ell_n_estim}
\end{equation}

Note that since we separated the parameters from random samples, 
the gradient of Eq.~\ref{eq:ell_n_estim} can be derived for individual 
terms with respect to $\Psi$ and $\Theta$, yielding an unbiased estimate of 
the gradient of Eq.~\ref{eq:ell_n}. 

Three options were used to perform DRF inference in Cutajar {\em et al.}~\cite{rfdnn17}, known as  PRIOR-FIXED, VAR-FIXED, and VAR-RESAMPLED. With PRIOR-FIXED,  the random spectra $\Omega$ is not inferred (for simplicity), but marginalized out 
from Eq.~\ref{eq:post_orig}). Then only the parameters $\Lambda$ are trained. This can be achieved by removing the KL term regarding $\Omega$ in Eq.~\ref{eq:elbo} and use $\Omega^{(s)}$ sampled from $P(\Omega|\Lambda)$ in 
Eq.~\ref{eq:ell_n_estim} instead of Eq.~\ref{eq:reparam_Omega}. 

With VAR-FIXED and VAR-RESAMPLED, $\Omega$ is inferred  in the posterior $q(\Omega)$ with the corresponding variational parameters $\Psi_\Omega$ (this is shown in our derivation above). The difference between the two VAR options is whether the random numbers $\{ e^{(s)}_{lij}, \tau^{(s)}_{lij}, \epsilon^{(s)}_{nj} \}_{l,i,j,n}$ are sampled once and fixed throughout the optimization (VAR-FIXED), or sampled 
at every iteration (VAR-RESAMPLED).


\subsection{Prediction}\label{sec:prediction}

Given a trained model, where the variational density $q(\cdot)$ and model parameters $\Theta$ are optimized, we predict the model's 
output and its uncertainty for an unseen test input $x_*$ as follows. 
Let $F_* = [f_1(x_*), \dots, f_{d_0}(x_*)]^\top$ be the output vector of 
the bottom GP layer on $x_*$ (also the input vector to the upper DRF), 
and $y_*$ the final target output of the model.
The posterior distribution for $y_*$ is approximated as,
\begin{equation}
\begin{split}
P(y_*| x_*, X, \overline{X}, Y, \Theta) \approx
    \int &P(y_*|G(F_*; W, \Omega, \theta_o), \theta_l) 
    \\\ &P(F_*|\overline{F}) 
    \ q(W, \Omega, \overline{F}) \ dW d\Omega \ d\overline{F}.
\label{eq:pred_ys}
\end{split}
\end{equation}

Although the last two integrands in Eq.~\ref{eq:pred_ys} are Gaussians, the 
first term is highly involved with integration variables, analytic solution 
is infeasible. Rather, we do the Monte-Carlo estimation similar to what we did 
in Section~\ref{sec:inference}. That is, after sampling $(W^{(s)}, \Omega^{(s)}, F^{(s)}_*)$ from the Gaussian of the last two integrands, we have the approximation of Eq.~\ref{eq:pred_ys} as:
\begin{equation}
\frac{1}{S} \sum_{s=1}^S 
    P(y_*|G(F^{(s)}_*; W^{(s)}, \Omega^{(s)}, \theta_o), \theta_l).
\label{eq:pred_ys_mc}
\end{equation}
Then we can represent the posterior distribution of $y_*$ by the samples 
$\{ y^{(t)}_* \}_{t=1}^T$ which are obtained by sampling from the mixture 
density defined in Eq.~\ref{eq:pred_ys_mc}. Namely, for each $t=1,\dots,T$, 
(i) select $s$ uniformly at random from $\{1,\dots,S\}$, then (ii) sample 
$y^{(t)}_* \sim P(y_*|G(F^{(s)}_*; W^{(s)}, \Omega^{(s)}, \theta_o), \theta_l)$ 
which requires a full feed-forward pass of the input $x_*$ through the 
GP-DRF model. Therefore, for a scalar target $y_*$, the posterior mean 
and variance can be estimated as:
\begin{eqnarray}
\mathbb{E}[y_*|x_*, X, \overline{X}, Y] &\approx& 
    \frac{1}{T} \sum_{t=1}^T y^{(t)}_* \ \ \ \ \ \ (:= \overline{y}_*) \\
\mathbb{V}(y_*|x_*, X, \overline{X}, Y) &\approx& 
    \frac{1}{T-1} \sum_{t=1}^T (y^{(t)}_*-\overline{y}_*)^2.
\end{eqnarray}

\section{Experiments}\label{sec:experiments}

\begin{table*}[t]\vspace{-3mm}
\centering
\caption{Dataset Statistics and Benchmark Results.}
\label{tab:statistics}
\tablestyle{4pt}{1.05}
\begin{tabular}{l|x{35}|x{35}|x{35}|x{35}|x{35}|x{35}|x{35}|x{35}}
\multirow{1}{*}{{\textbf{}}}  & 
\multicolumn{1}{c|}{\textbf{POWERPLANT}} &\multicolumn{1}{c|}{\textbf{PROTEIN}} &  \multicolumn{1}{c|}{\textbf{SPAM}}& 
\multicolumn{1}{c|}{\textbf{EEG}}& 
\multicolumn{1}{c|}{\textbf{MNIST}} & 
\multicolumn{1}{c|}{\textbf{MUSIC}} & 
\multicolumn{1}{c|}{\textbf{REUTERS}}& 
\multicolumn{1}{c}{\textbf{SCOP}}   \\ \shline
\multicolumn{9}{c}{\textbf{Dataset Statistics}}   \\ \shline
\# Train        & 9469 & 45515 & 4532& 14857& 60000 &900& 8084 & 2575    \\
\# Test & 99 & 215 & 69& 123& 10000 & 100 & 899 & 287  \\
\# Classes &  1 & 1 & 2 & 2&10 & 10& 2& 7 \\\shline
\multicolumn{9}{c}{\textbf{Benchmark Results} (Lower is better; best method in bold)}   \\ \shline
 GP       & 0.207 & 0.737 &0.043 &0.114 & 0.059 &0.350& 0.050 &0.301     \\
DRF & 0.201 &  \textcolor{black}{\textbf{0.613}} & 0.029& 0.106&  0.045&  0.660&  0.020&   0.381\\
 GP-DRF (Ours) &  \textcolor{black}{\textbf{0.194}} &0.652  & \textcolor{black}{\textbf{0.001}} & \textcolor{black}{\textbf{0.016}} & \textcolor{black}{\textbf{0.033}} & \textcolor{black}{\textbf{0.300}} & \textcolor{black}{\textbf{0.017}} & \textcolor{black}{\textbf{0.286}}  \\
\end{tabular}\vspace{-1mm}
\end{table*}

\begin{figure*}[t]
\centering
\begin{tabular}{lll}

\begin{subfigure}[b]{0.31\textwidth}
\includegraphics[scale=0.445]{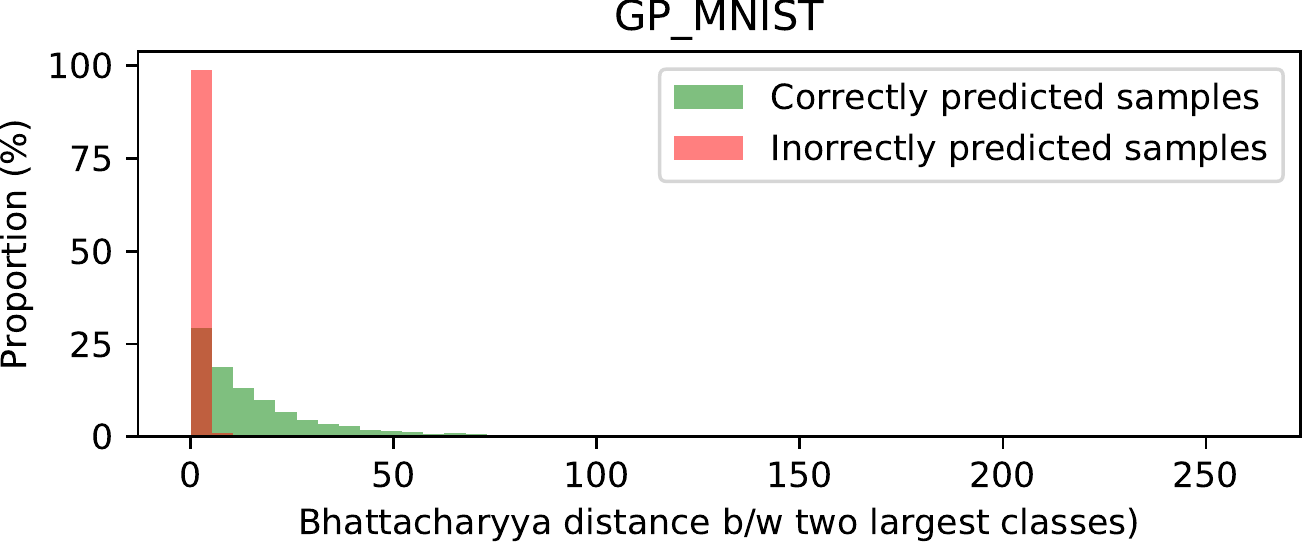}
\caption{GP on MNIST.}
\end{subfigure}
&
\begin{subfigure}[b]{0.31\textwidth}
\includegraphics[scale=0.445]{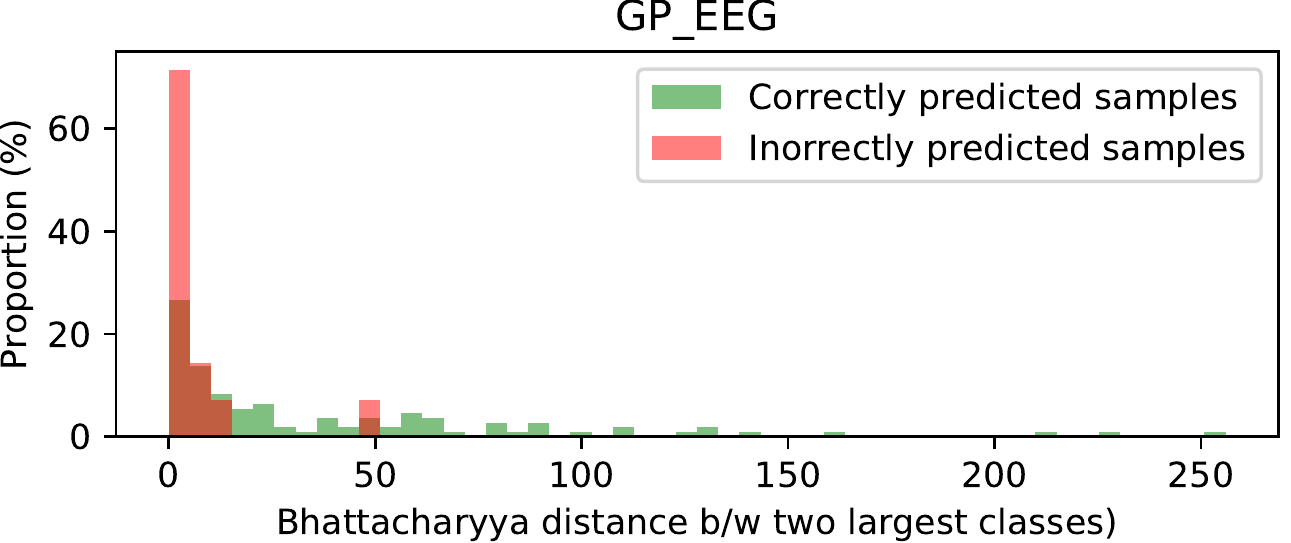}
\caption{GP on EEG.}
\end{subfigure}
&
\begin{subfigure}[b]{0.31\textwidth}
\includegraphics[scale=0.445]{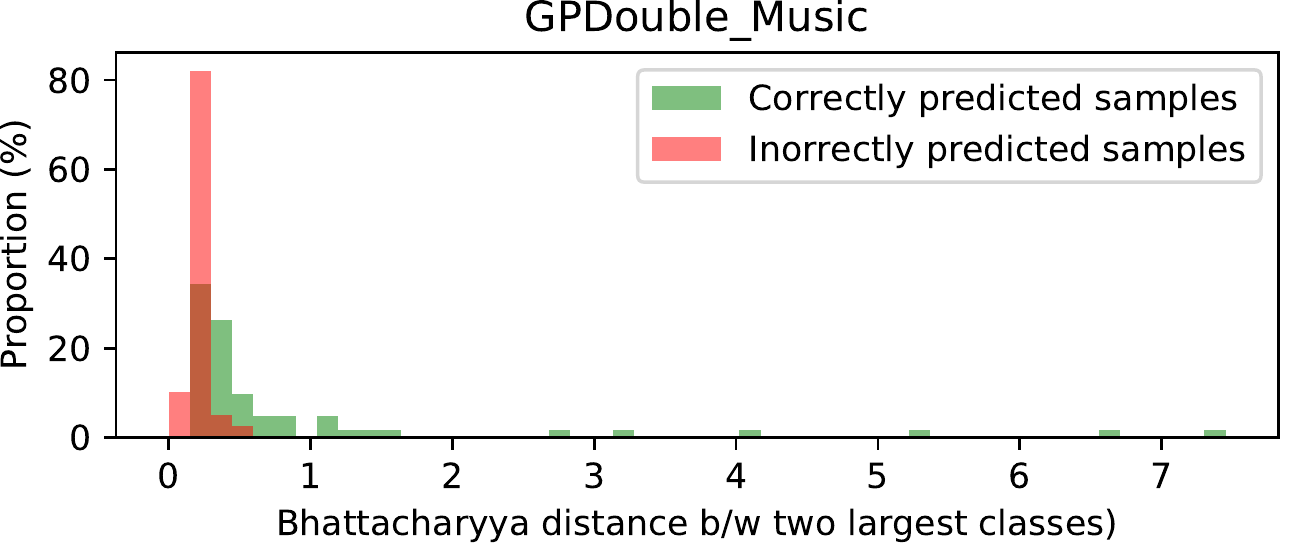}
\caption{GP on Music.}
\end{subfigure}
\\

\begin{subfigure}[b]{0.31\textwidth}
\includegraphics[scale=0.445]{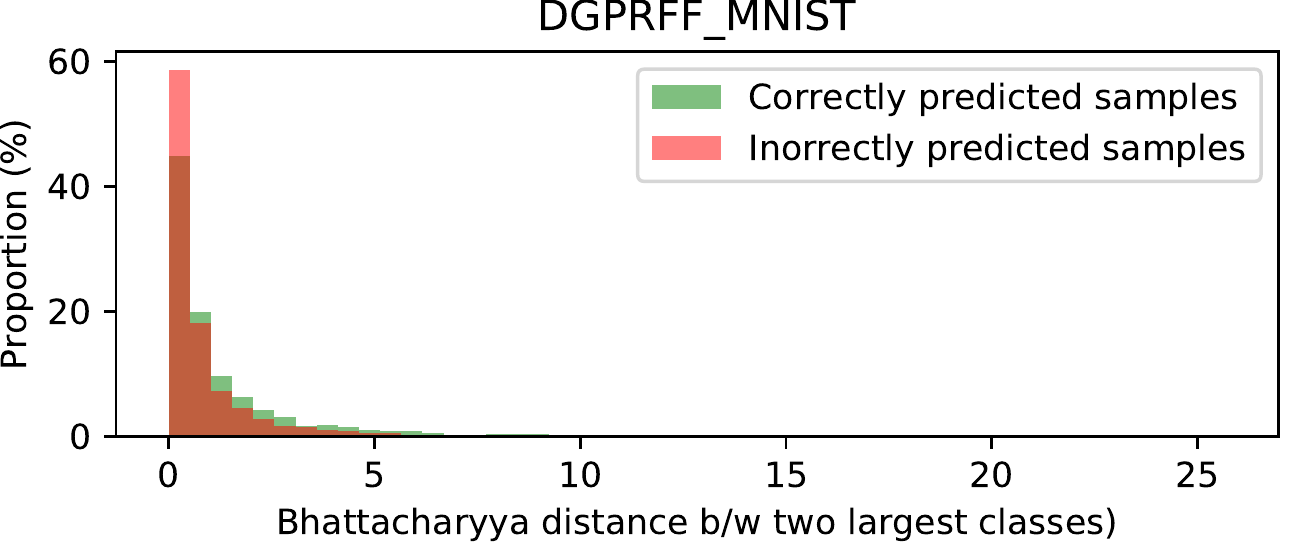}
\caption{DRF on MNIST.}
\end{subfigure}
&
\begin{subfigure}[b]{0.31\textwidth}
\includegraphics[scale=0.445]{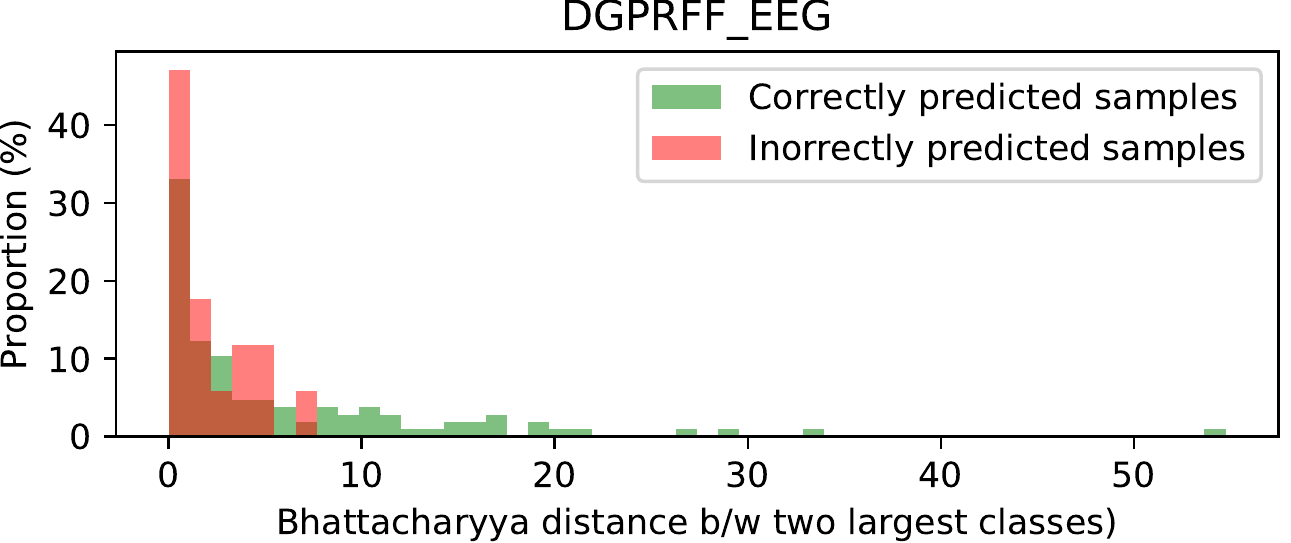}
\caption{DRF on EEG.}
\end{subfigure}
&
\begin{subfigure}[b]{0.31\textwidth}
\includegraphics[scale=0.445]{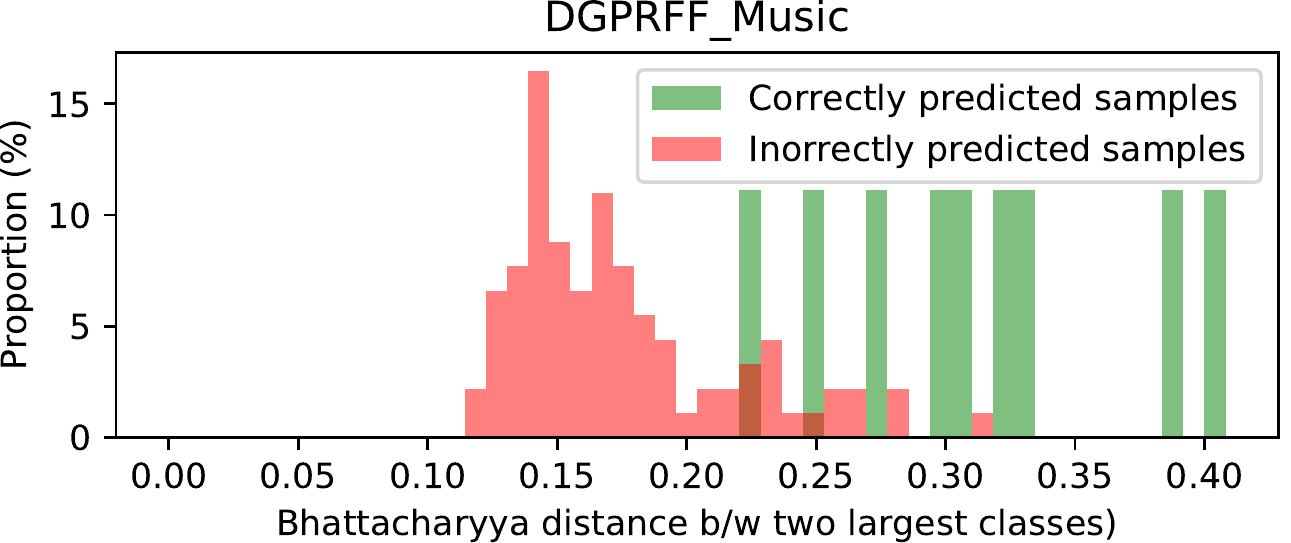}
\caption{DRF on Music.}
\end{subfigure}
\\

\begin{subfigure}[b]{0.31\textwidth}
\includegraphics[scale=0.445]{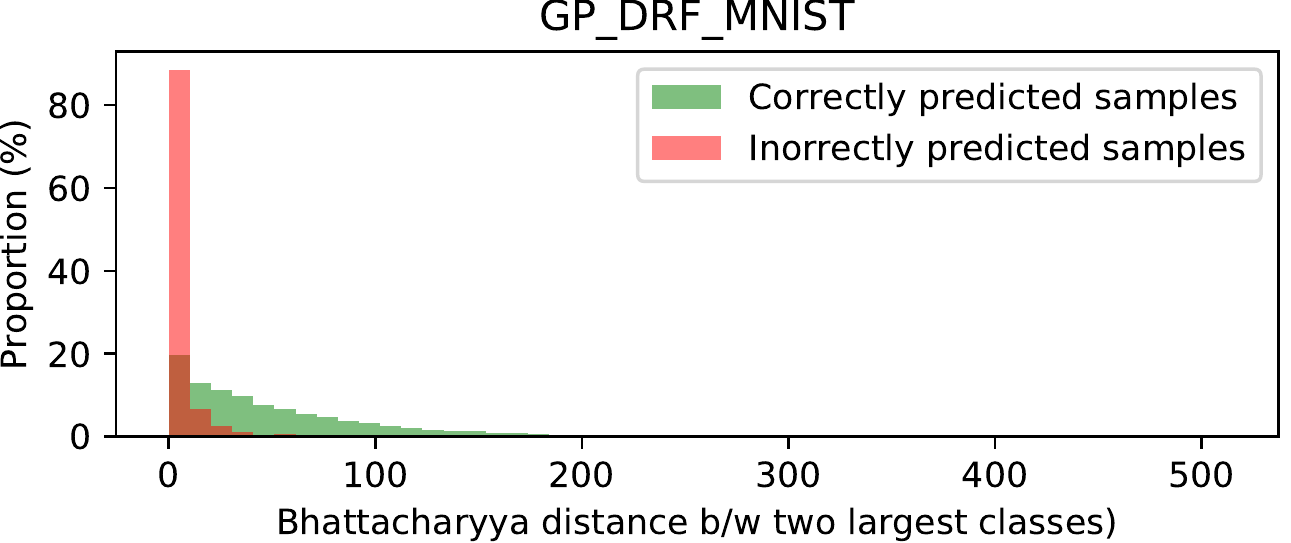}
\caption{GP-DRF on MNIST.}
\end{subfigure}
&
\begin{subfigure}[b]{0.31\textwidth}
\includegraphics[scale=0.445]{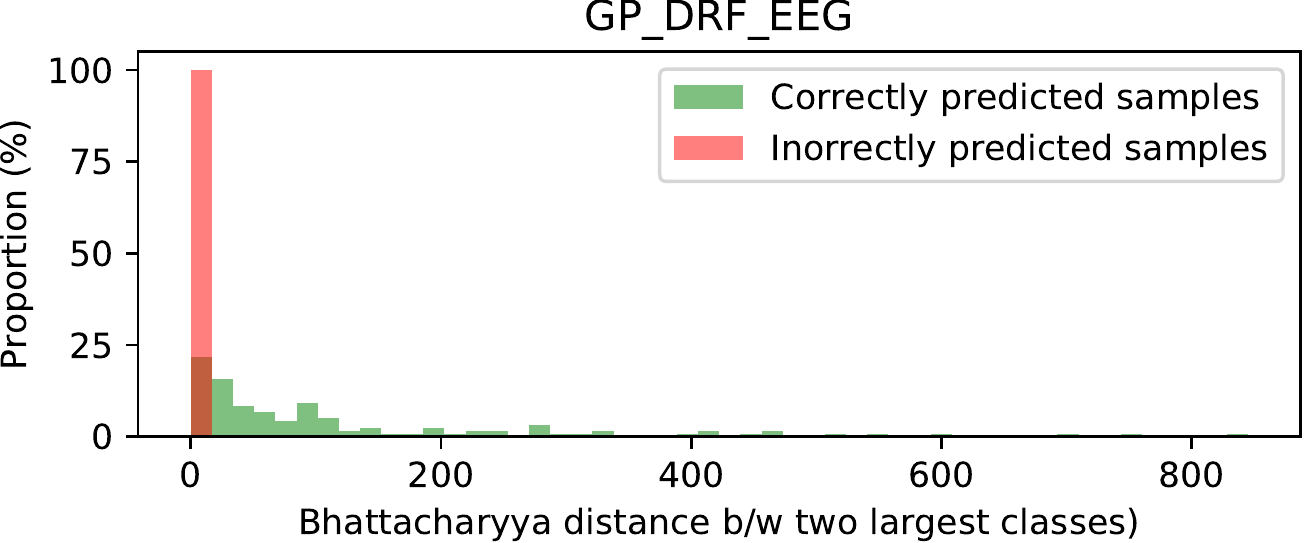}
\caption{GP-DRF on EEG.}
\end{subfigure}
&
\begin{subfigure}[b]{0.31\textwidth}
\includegraphics[scale=0.445]{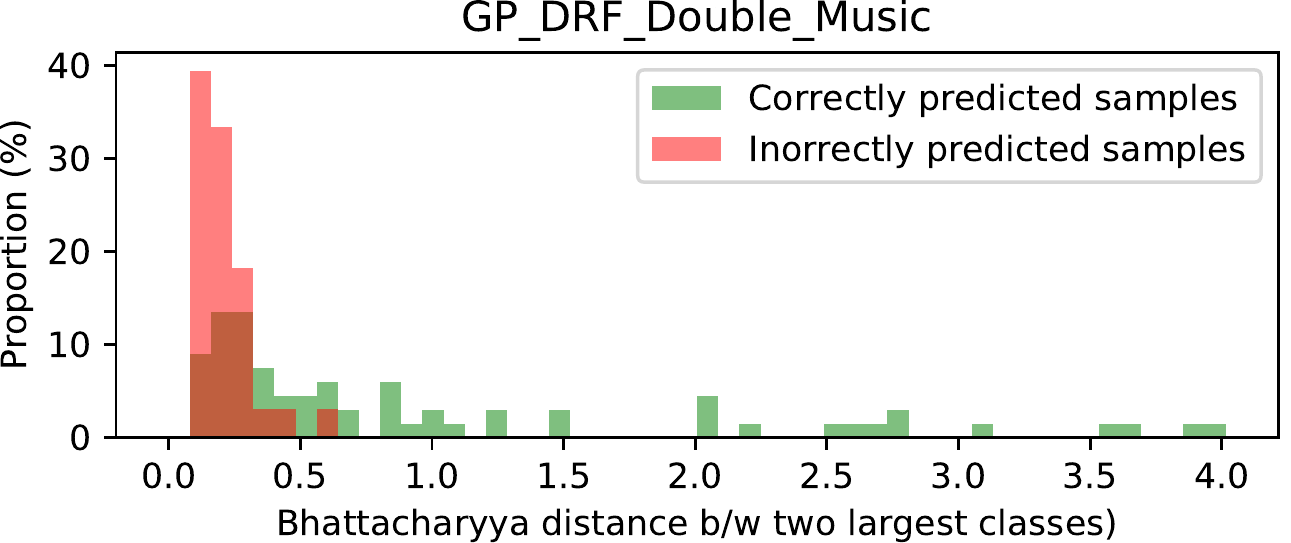}
\caption{GP-DRF on Music.}
\end{subfigure}

\end{tabular}

\caption{Bhattacharyya distances for correctly labeled and misclassified samples evaluated for three competing models, GP, DRF, and our GP-DRF, on three datasets, MNIST, EEG, and Music. }
\label{fig:Bhattacharyya}
\end{figure*}

\begin{table*}[t]
\centering
\caption{Comparison with respect to the average Battaacharya distance for correctly labeled (higher is better) and misclassified samples (lower is better). The best scores are boldfaced.}
\label{tab:battaacharya}
\begin{tabular}{l|x{55}x{55}|x{55}x{55}|x{55}x{55}}
\multirow{2}{*}{{\textbf{Model}}}  & 
\multicolumn{2}{c|}{\textbf{EEG}}& 
\multicolumn{2}{c|}{\textbf{MNIST}} & 
\multicolumn{2}{c}{\textbf{Music}}   \\ 
 & \textbf{correctly} $\uparrow$&  \textbf{misclassified} $\downarrow$ &  \textbf{correctly} $\uparrow$&  \textbf{misclassified} $\downarrow$ &  \textbf{correctly} $\uparrow$&  \textbf{misclassified} $\downarrow$  \\\shline
 GP & 37.41&  6.37 & 18.85& 0.79 & 0.89& \textcolor{black}{\textbf{0.21}}  \\
  DRF & 6.16&\textcolor{black}{\textbf{2.19}}&0.31&\textcolor{black}{\textbf{0.18}}&\textcolor{black}{\textbf{1.40}}&0.97 \\
  GP-DRF (Ours) & \textcolor{black}{\textbf{110.52}} &5.45&\textcolor{black}{\textbf{50.17}} &4.65&0.99&\textcolor{black}{\textbf{0.21}} \\
 \\
\end{tabular}\vspace{-.5em}
\vspace{-3mm}
\end{table*}

\subsection{Experimental Setup}
To showcase the efficacy of GP-DRF, we evaluate it on several datasets, grouped into 3 tasks: (1) fixed-sized input classification task, which includes MNIST\cite{lecun1998gradient}, EEG, and SPAM~\cite{Dua:2017}; (2) fixed-sized input regression task, POWERPLANT, and PROTEIN~\cite{Dua:2017}; and (3) variable-sized (sequence) input classification task, which includes MUSIC~\cite{li2003comparative} (a music genre dataset for multi-class genre prediction), REUTERS\footnote{Available at,\\ \url{http://www.daviddlewis.com/resources/testcollections/reuters21578/}} (a text dataset for text categorization, and SCOP~\cite{lo2000scop} (a protein sequence dataset for protein fold recognition). Their statistics are described in Table~\ref{tab:statistics}. The evaluation metric for classification datasets is the mean number of misclassifications (error rate), and, for regression datasets, the root mean square error (RMSE).

We compare GP-DRF against two baselines: (1) GP, and (2) DRF. GP is a Gaussian process based model with the same architecture as the first layer of GP-DRF. Each target class is associated with a Gaussian process, and is trained using variational inference as described in Hensman {\em et al.} \cite{hensman2015scalable}. DRF represents the same architecture as the DRF component of GP-DRF, and we train it using the procedure in Cutajar {\em et al.}~\cite{rfdnn17}. For the sequence datasets, the models use the double (1.5) kernel features as described in Kuksa {\em et al.}~\cite{kuksa2008kernel}, and the ARD kernel (as described in Cutajar {\em et al.}~\cite{rfdnn17}) features for the rest of the datasets. For the Gaussian processes, each kernel feature $K_i(x, y)$ is associated with two trainable parameters: (1) $\alpha_i$ which scales the output as $\alpha_i \cdot K_i(x, y)$ , and (2) $\sigma_i$ which is a parameter within the kernel function.

\subsection{Implementation Details}
We run the ADAM~\cite{kingma2014adam} optimizer for 1000 epochs with learning rate $1\times10^{-5}$. L2 penalty is added to all parameters with the coefficient $5\times10^{-4}$. For GP and GP-DRF, the number of inducing points is 200. At each iteration, a single example is selected uniformly at random from the training set and 100 MCMC samples are collected from each random variable. Each model uses the Gaussian likelihood for regression and the softmax likelihood for classification problems.

\subsection{Comparison to GP and DRF Models}
Table~\ref{tab:statistics} shows that GP-DRF consistently outperforms GP and DRF on all eight datasets. Further, GP-DRF reduces the error rate of ``Double-(1,5) (MFCC)"~\cite{kuksa2008kernel} by $7.7\%$  on the Music dataset while having uncertainty quantification. This suggests that combining exact and approximate approaches to computing kernel features, and leveraging deep structures can be useful.

\subsection{Bhattacharyya Distance Benchmark}
The Bhattacharyya distance~\cite{bhattacharyya1946measure} is a widely used measure within the research community~\cite{schweppe1967state, choi2003feature, michailovich2007image}. It can be used to measure the separability of classes in classification. It is more reliable than the Mahalanobis distance~\cite{mahalanobis1936generalized} as the Bhattacharyya distance grows depending on the difference between the means of the classes as well as their standard deviations, rather than just the means.

In this setup, we perform uncertainty analysis on our models by computing the distance between ``the two most confident class posterior distributions" with respect to the Bhattacharaya measure. For a K-way classification task, the certainty is
\begin{equation}
\begin{split}
    D(F_*(x), F_+(x)) = &\frac{1}{4}\ln{\Big(\frac{1}{4}\Big(\frac{\sigma_*}{\sigma_+}
                        +\frac{\sigma_+}{\sigma_*}+2}\Big)\Big) \\&+ \frac{1}{4}\Big(\frac{(\mu_*-\mu_+)^2}{\sigma_*+\sigma_+}\Big),
\end{split}
\end{equation}
where $F_*(x) = \mathcal{N}(\mu_*, \sigma_*)$ and is the distribution over the posterior samples obtained for the test example's most confident predicted class; and $F_+(x) = \mathcal{N}(\mu_+, \sigma_+)$ represents that of the test example's second most confident predicted class. This is the notion of "margin" in class prediction, where the larger distance suggests the model is more certain about its prediction.

Table~\ref{tab:battaacharya} shows the average Bhattacharyya distances for the correctly, $D_c$, and misclassified samples, $D_m$, on the three datasets. We see that GP-DRF has the largest discrepancy between $D_c$ and  $D_m$, suggesting it is significantly more confident than competing models when making correct prediction. 

The histograms in Figure~\ref{fig:Bhattacharyya} show the Bhattacharyya distances for each sample in the test set for the correctly classified (shown as green bars) and the misclassified samples (red bars). The histograms further justify GP-DRF's efficacy, as it offers higher certainty compared to GP and DRF when it correctly classifies a test example. This implies that the quantitative measure of prediction uncertainty, derived from our Bayesian model, can be used as an accurate gauge of the quality of prediction.


\section{Conclusion}
We proposed GP-DRF, a novel deep Gaussian process, which defines a powerful Bayesian model that is scalable, can deal with sequential inputs, provides uncertainty estimates, and achieves superior performance compared to its counterparts. It combines the non-parametric structure of Gaussian processes in its first layer and the parametric approximation of Gaussian processes in the rest of the network. GP-DRF consistently outperforms the GP and the DRF models on several benchmarks. GP-DRF can also provide better certainty estimates, quantified by the Battacharaya distance. In our future work, we will explore other structured, variable-size data domains, including, graph and language data.

\section{Acknowledgements}
We would like to thank the anonymous referees for their constructive comments and suggestions. Issam Laradji was funded by the UBC Four-Year Doctoral Fellowships (4YF).

\bibliographystyle{unsrt}
\bibliography{refs}

\end{document}